\documentclass[fleqn,10pt]{wlscirep}
\usepackage[utf8]{inputenc}
\usepackage[T1]{fontenc}

\title{Scientific Reports Title to see here}

\usepackage{subfig}
\usepackage{multirow}
\usepackage{comment}

\def\eg{\emph{e.g., }}
\newcommand{\bfsection}[1]{\vspace*{0.1cm}\noindent\textbf{#1.}}

\title{Aphid Cluster Recognition and Detection in the Wild Using Deep Learning Models}

\author[1]{Tianxiao Zhang}
\author[1]{Kaidong Li}
\author[1]{Xiangyu Chen}
\author[1]{Cuncong Zhong}
\author[1]{Bo Luo}
\author[2]{Ivan Grijalva}
\author[2]{Brian McCornack}
\author[3]{Daniel Flippo}
\author[3]{Ajay Sharda}
\author[4,*]{Guanghui Wang}
\affil[1]{Department of Electrical Engineering and Computer Science, University of Kansas, Lawrence, KS 66045, USA}
\affil[2]{Department of Entomology, Kansas State University, Manhattan, KS 66506, USA}
\affil[3]{Department of Biological and Agricultural Engineering, Kansas State University, Manhattan, KS 66506, USA}
\affil[4]{Department of Computer Science, Toronto Metropolitan University, Toronto, ON M5B 2K3, Canada}

\affil[*]{Corresponding author: Guanghui Wang (wangcs@torontomu.ca)}

\begin{abstract}
Aphid infestation poses a significant threat to crop production, rural communities, and global food security. While chemical pest control is crucial for maximizing yields, applying chemicals across entire fields is both environmentally unsustainable and costly. Hence, precise localization and management of aphids are essential for targeted pesticide application. The paper primarily focuses on using deep learning models for detecting aphid clusters.
We propose a novel approach for estimating infection levels by detecting aphid clusters. To facilitate this research, we have captured a large-scale dataset from sorghum fields, manually selected 5,447 images containing aphids, and annotated each individual aphid cluster within these images. To facilitate the use of machine learning models, we further process the images by cropping them into patches, resulting in a labeled dataset comprising 151,380 image patches. Then, we implemented and compared the performance of four state-of-the-art object detection models (VFNet, GFLV2, PAA, and ATSS) on the aphid dataset. Extensive experimental results show that all models yield stable similar performance in terms of average precision and recall. We then propose to merge close neighboring clusters and remove tiny clusters caused by cropping, and the performance is further boosted by around 17\%. The study demonstrates the feasibility of automatically detecting and managing insects using machine learning models. The labeled dataset will be made openly available to the research community. 
\end{abstract}
\begin{document}

\flushbottom
\maketitle
%
%
\thispagestyle{empty}

\section{Introduction}

According to the Food and Agriculture Organization of the United Nations \cite{fao2023}, pests annually destroy up to 40\% of global crops, causing a specific loss of around \$70 billion due to invasive insects. In modern agriculture, pesticides are extensively used to control insect populations, amounting to approximately 2 million tons of pesticide usage worldwide each year \cite{sharma2019worldwide}. However, the widespread use of pesticides contributes to environmental pollution and poses significant threats to wildlife and human beings  \cite{sharma2019worldwide}. Therefore, efficient management of pesticide applications is essential for mitigating environmental impact and enhancing farmer profitability.

Due to practical constraints such as time, labor, and lack of automated resources or technologies, the current management strategy is predominantly guided by a "whole field approach." Farmers typically wait for the infestation to reach a treatment threshold before spraying the entire field using a linear array of nozzles. This approach leads to an excessive, uniform, and continuous spray pattern, with all plants receiving the same treatment irrespective of their individual infection levels, while pest incidence and severity are typically only fractionally present in a field and spread spatially. Consequently, the current practice only guarantees a small fraction of areas receives a justified amount of pesticide. Some areas may suffer from pest damage and loss of yields due to delayed management, while other areas may receive a superfluous spray application when there is no pest presence. 


Timing the incidence and severity of insect infestations presents a significant challenge in effectively managing crop damage and maximizing profits. With the rapid advancement of robotics and artificial intelligence (AI), we aim to develop a robotic system capable of regularly scouting the field, utilizing AI to identify insect infestations, and precisely spraying affected areas through onboard spray nozzles. This intelligent system will selectively apply pesticides only to plants with critical infestations, minimizing pesticide usage on healthy plants and the surrounding soil. As a result, we can reduce input costs, increase the potential yield, and minimize the environmental impact of pesticides. The paper primarily focuses on a crucial component of the system: automatically detecting insects using computer vision technology and assessing the performance of various deep learning models for detecting aphid clusters.

Object detection and recognition are crucial components in agricultural robotics, and detecting small insects like aphids can be particularly challenging. The use of Convolutional Neural Networks (CNNs) for object detection and recognition was first proposed in \cite{girshick2014rich}. In 2012, the breakthrough success of deep CNNs in the ImageNet challenge marked a new era of this technique. Since then, CNN models have been widely used in various applications, including  biology \cite{spiesman2021assessing}, medical image analysis \cite{li2021colonoscopy}, object detection and tracking \cite{zhang2020real}, etc. In recent years, "end-to-end" training models, such as SSD \cite{liu2016ssd}, RetinaNet \cite{lin2017focal}, and FCOS \cite{tian2019fcos}, have been extensively studied. Object detection models are more suitable for detecting insects like aphids as they require both localization and classification. These models can be categorized into two-stage and one-stage approaches. Two-stage methods, such as Faster R-CNN \cite{ren2015faster} and FPN \cite{lin2017feature}, first use Region Proposal Networks (RPNs) \cite{ren2015faster} to select high-quality pre-defined anchor boxes as candidate objects, then refine the selected candidates in the second stage for classification and bounding box regression. One-stage methods, such as RetinaNet \cite{lin2017focal} and FCOS \cite{tian2019fcos}, directly classify and refine pre-defined anchor boxes or anchor points. Most recent detection models are one-stage methods due to their fast inference and relatively high accuracy.

However, even with state-of-the-art detection models, accurately locating individual aphids remains a challenge due to their small size. To address this challenge, TD-Net \cite{teng2022td} introduces a T-FPN (Transformer feature pyramid network) and a multi-resolution training method, while ZF-Net \cite{zeiler2014visualizing} and RPN \cite{ren2015faster} are employed in \cite{li2019automatic} for aphid detection on leaves. Another model, Coarse Convolutional Neural Network (CCNN) \cite{li2019coarse}, is designed to detect small aphid clusters, and a Fine Convolutional Neural Network (FCNN) is used to refine the clique and detect individual aphids. 
Most existing aphid detection models primarily focus on detecting individual aphids, but their performance on real aphid images is not satisfactory. These models are often trained on idealized aphid images, making it challenging for them to detect aphids accurately in real-world scenarios where aphids tend to cluster together on leaves. The task of accurately delineating and detecting densely-packed aphids individually is nearly impossible. Additionally, domain shifts caused by variations in illumination and shades across different images can significantly impact the accuracy of CNN models in detecting tiny aphids \cite{yang2022unsupervised, zhang2021six}.

In order to train the machine learning models to detect aphids, we captured a large collection of images in the sorghum fields of the State of Kansas during the aphids growing season. Then, we carefully examined all images and selected and labeled 5,447 images that exhibited signs of aphid infestation. In this study, we manually annotate the outline of the aphid clusters rather than labeling individual aphids. The size of these clusters can indicate the severity of the aphid infestation \cite{raiyan2023}. Since the aphid clusters are much larger than individual aphids, most standard object detection models can effectively localize these clusters. Using the annotated dataset, we implemented and compared the performance of four state-of-the-art detection models. Despite creating bounding boxes around aphid clusters, accurately dividing these clusters can still pose challenges that impact the detectors' performance. To address this, we merged closely located clusters into larger ones and adjusted the coordinates of the bounding boxes accordingly, forming a minimum bounding box around the merged clusters.

Our experiments on the generated dataset demonstrate that these models can accurately estimate the extent of aphid infestation from real-world images, providing valuable information for farmers to make timely decisions regarding pest control. The main contributions of this study are summarized below.

\begin{itemize}
    \item We have collected a large collection of real-world images from sorghum fields and manually selected and labeled over 5,000 aphid-contaminated images. This is the first dataset using real-world images and employing cluster annotations instead of individual aphids.
    
    \item We evaluated the performance of four state-of-the-art detection models on the generated dataset, and we observed improved accuracy by merging nearby clusters and excluding tiny clusters from training.
    
    \item The existing detection models could be directly utilized for our dataset and the trained detectors could be employed in real-world fields to detect those aphid clusters. There is no need for sophisticated designs for detecting individual aphids, which makes it more feasible for real-world scenarios. 
\end{itemize}

Part results of this paper have been published at AAAI 2023 Workshop \cite{zhang2023new}. This is a substantial extension of the Workshop paper with more details of the dataset creation, deep learning models for object detection, experimental results, and analyses. The created dataset can be freely accessed by the research community from Harvard Dataverse at {https://doi.org/10.7910/DVN/N3YJXG}.

\section{Dataset Generation}

\subsection{Aphid Dataset}

 Instead of having a much closer view of the aphids and generating the aphid labels individually for the existing aphid datasets \cite{wu2019ip102}, we created the aphid dataset that contains images that have a relatively far view of the aphids in the real-world fields and the aphids are labeled by clusters instead of individuals since the aphids are frequently densely clustered together. The images we have taken and collected are shown in Fig~\ref{fig:samples}. It provides ground truth for both detection and segmentation. Our generated aphid dataset is more applicable to the elimination of the aphids if robots or intelligent cars are utilized. 
\subsection{Data Collection}

\begin{figure*}
\centering
    \subfloat[View 1]{
        \includegraphics[width=.33\columnwidth]{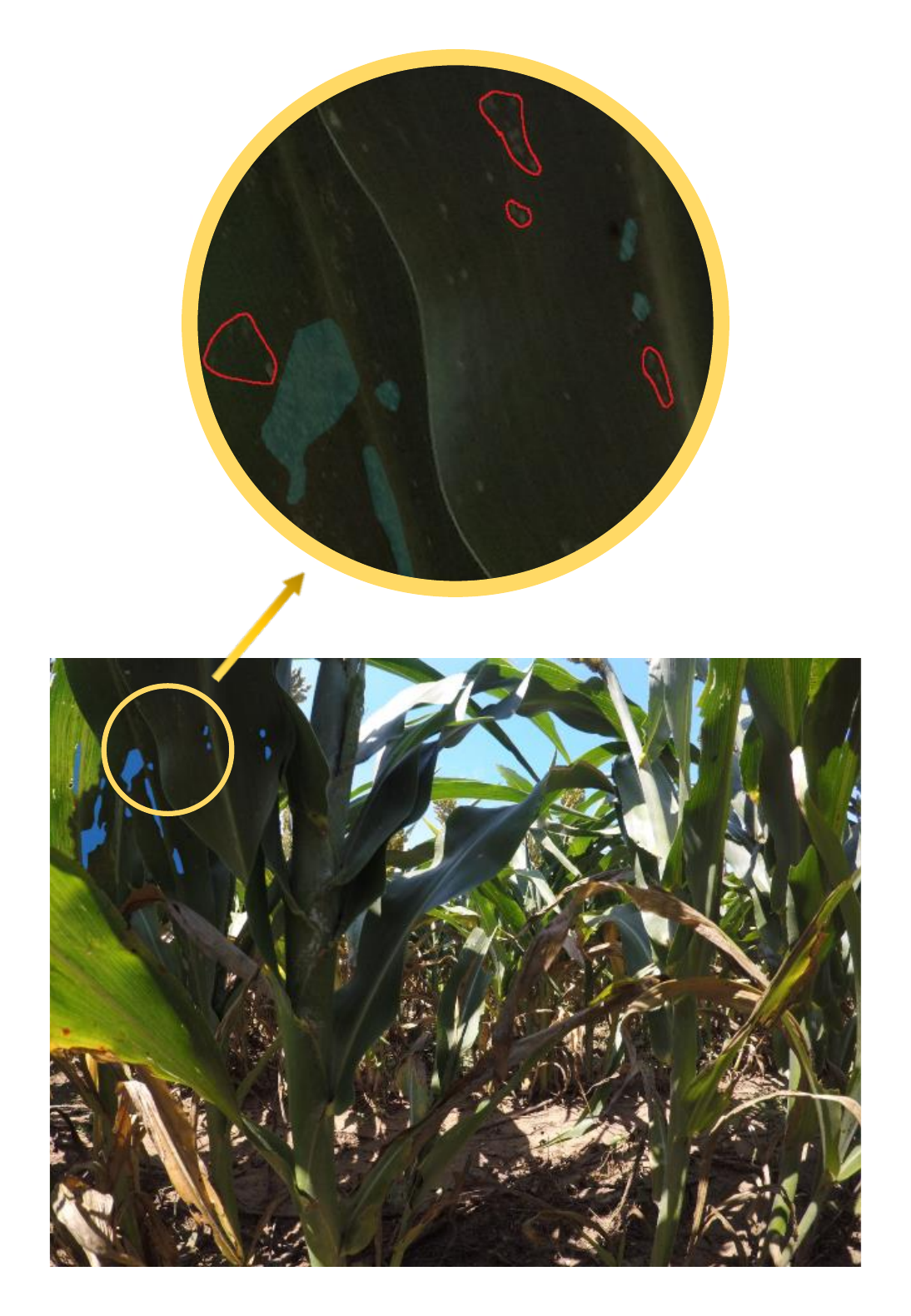} 
        \label{plots:tv}
    } 
    \subfloat[View 2]{
        \includegraphics[width=.33\columnwidth]{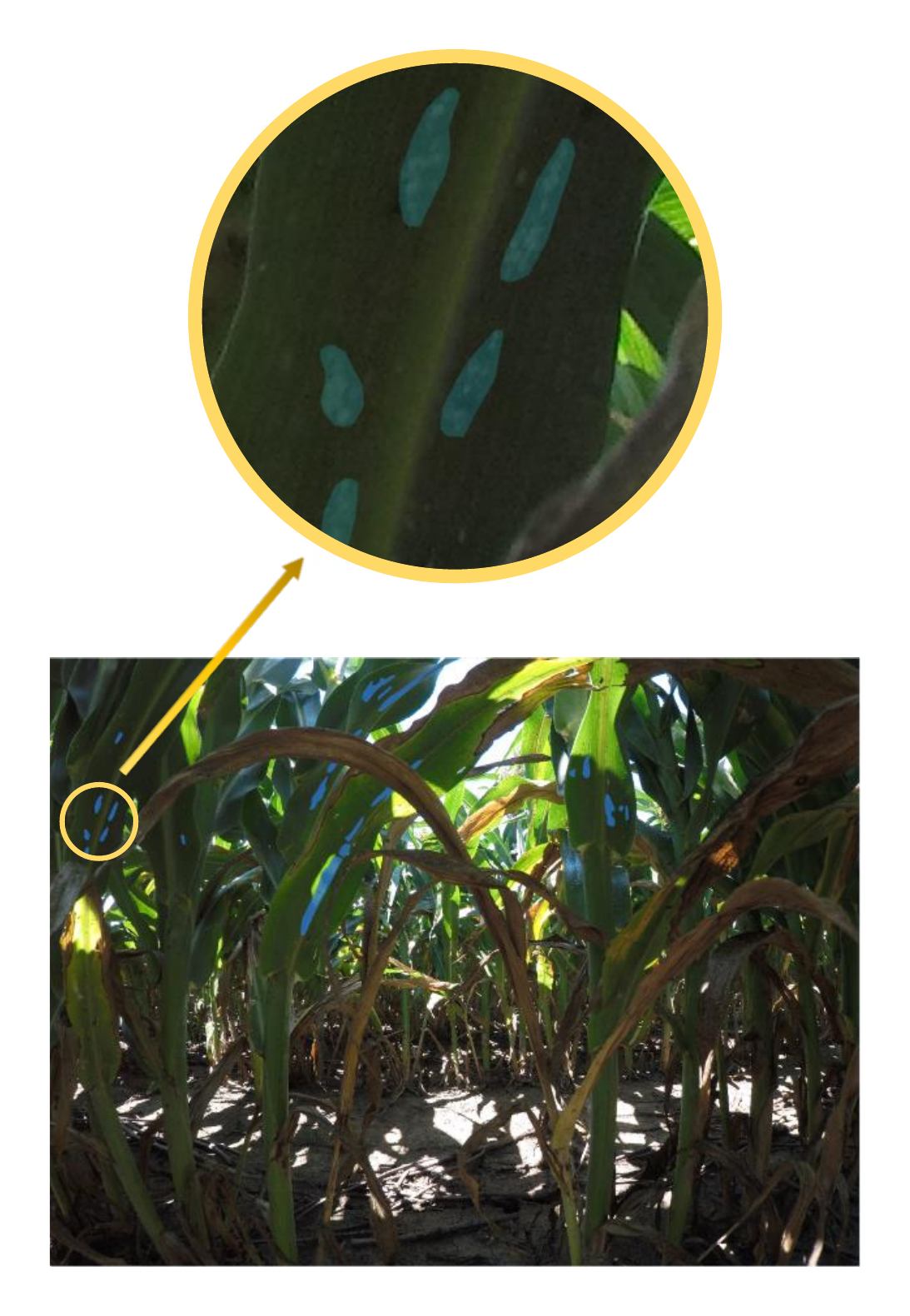} 
        \label{plots:mv}
    }
    \subfloat[View 3]{
        \includegraphics[width=.33\columnwidth]{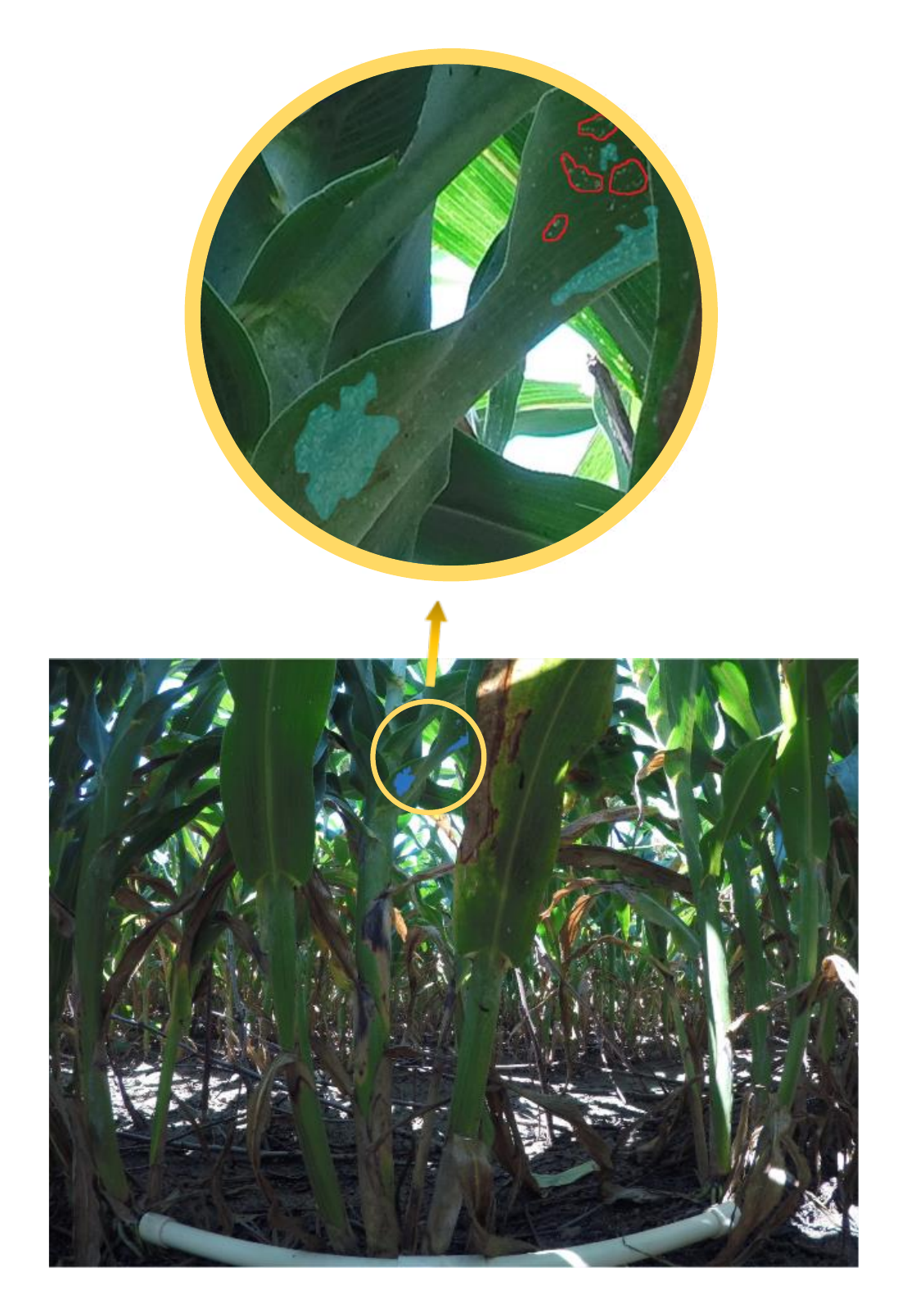} 
        \label{plots:bv}
    }
	\caption{{Original images with annotations}. The light blue areas represent an annotated aphid cluster. Aphid clusters are mostly tiny compared to the original image size. In (a) and (c), the areas circled by red lines represent clusters we do not need to annotate since those areas only contain very few sparsely distributed aphids. The criteria is an aphid cluster should have more than 6 closely located aphids.}
	\label{fig:samples}
\end{figure*}

To collect the images, we developed an imaging rig equipped with three GoPro Hero 5 cameras. This rig allows us to capture images of canopy leaves at three different heights, corresponding to view 1, view 2, and view 3 respectively. By doing so, we are able to enhance the dataset by capturing aphid clusters from various perspectives at multiple height levels. In Fig~\ref{fig:samples}, we present three annotated sample images representing the three views. Using this device, we extensively gathered a large collection of images from sorghum fields located in both Northern and Southern regions of the State of Kansas. These images were captured specifically in areas where aphid infections were identified. However, it is important to note that the majority of the images are actually aphid-free.

To annotate the dataset, we sought guidance from aphid experts and trained a team of eight research assistants to examine and label the dataset. We carefully examined all the images and eliminated those without aphids. Consequently, we successfully selected 5,447 images that contain an adequate number of aphids. The distribution of these images among the three views is illustrated in Fig~\ref{plots:compo}.
    
\subsection{Data Labeling}
The aphid clusters present in the selected images were meticulously annotated by professionally trained researchers using Labelbox, a tool specifically designed for data labeling \cite{labelbox}. The annotation process involved two main steps: creating segmentation masks for each image and generating detection bounding boxes based on these masks. 
These annotations provide valuable information for accurately identifying and studying the aphid populations within the images.

\bfsection{Aphid Cluster Definition} 
In the field, aphid clusters can exhibit various patterns, including low density, high density, and varying degrees of sparsity, as illustrated in Fig~\ref{fig:samples}. When it comes to labeling each individual aphid, irrespective of its density, a considerable amount of time and resources would be required, potentially wasting efforts on areas without a significant threat. Conversely, setting the threshold too high could lead to overlooking areas with substantial aphid infestations, resulting in potential financial losses.

To address this challenge and ensure consistency and productivity and avoid ambiguities, we consulted agricultural experts to establish a pragmatic definition of an aphid cluster. After careful consideration, we defined an aphid cluster as "an area containing six or more closely located aphids." This threshold strikes a balance between the need to accurately identify aphid clusters and the practical limitations of time and resources. Further illustration and interpretation of this threshold are shown in Fig~\ref{plots:tv} and \ref{plots:bv}, respectively. This definition ensures that areas with a critical level of aphid infestation are not overlooked while still optimizing the allocation of resources.

\begin{figure*}
\centering
    \subfloat[Dataset Composition]{
        \includegraphics[height=4.8cm]{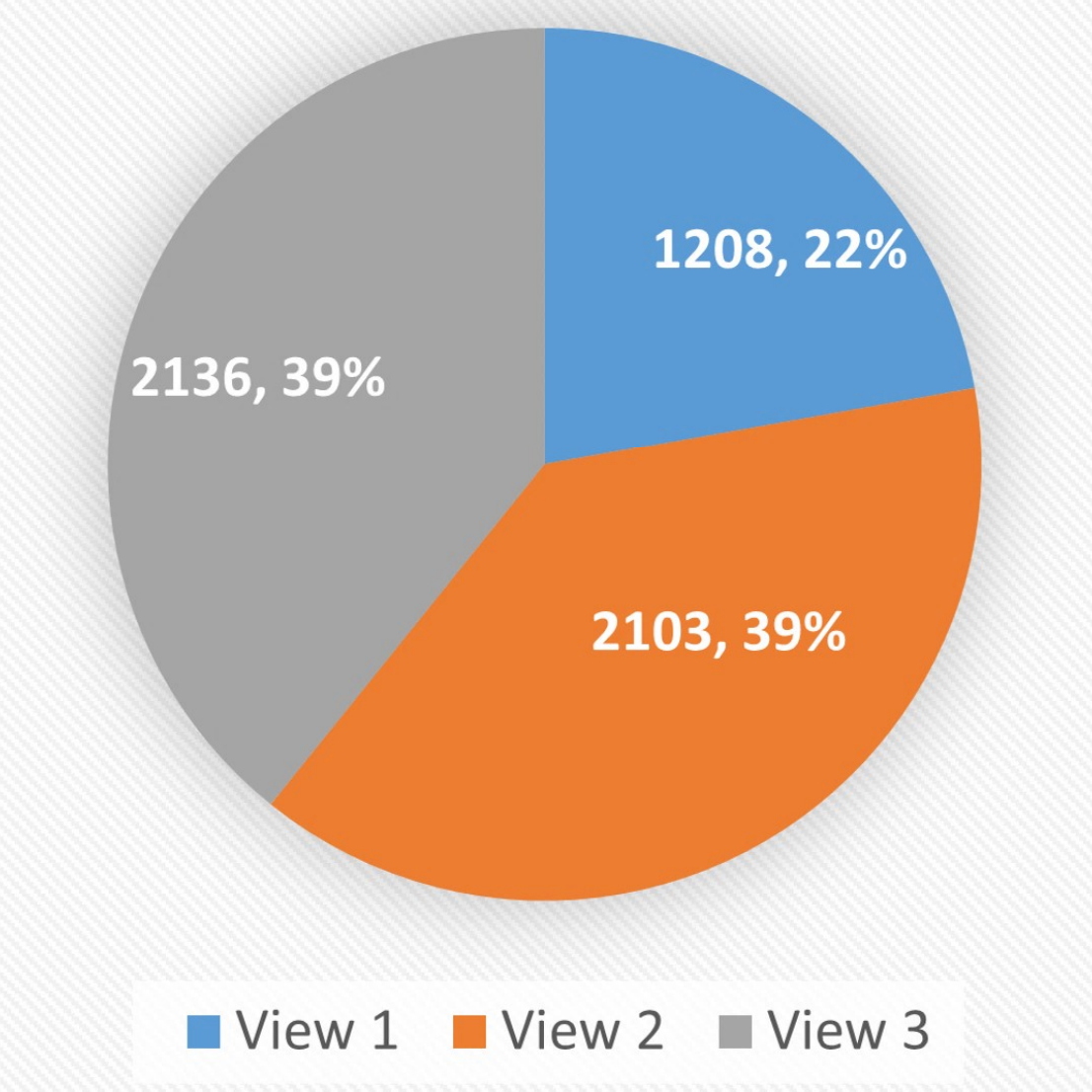}
        \label{plots:compo}
    }
    \subfloat[Mask Size Distribution]{
        \includegraphics[height=4.8cm]{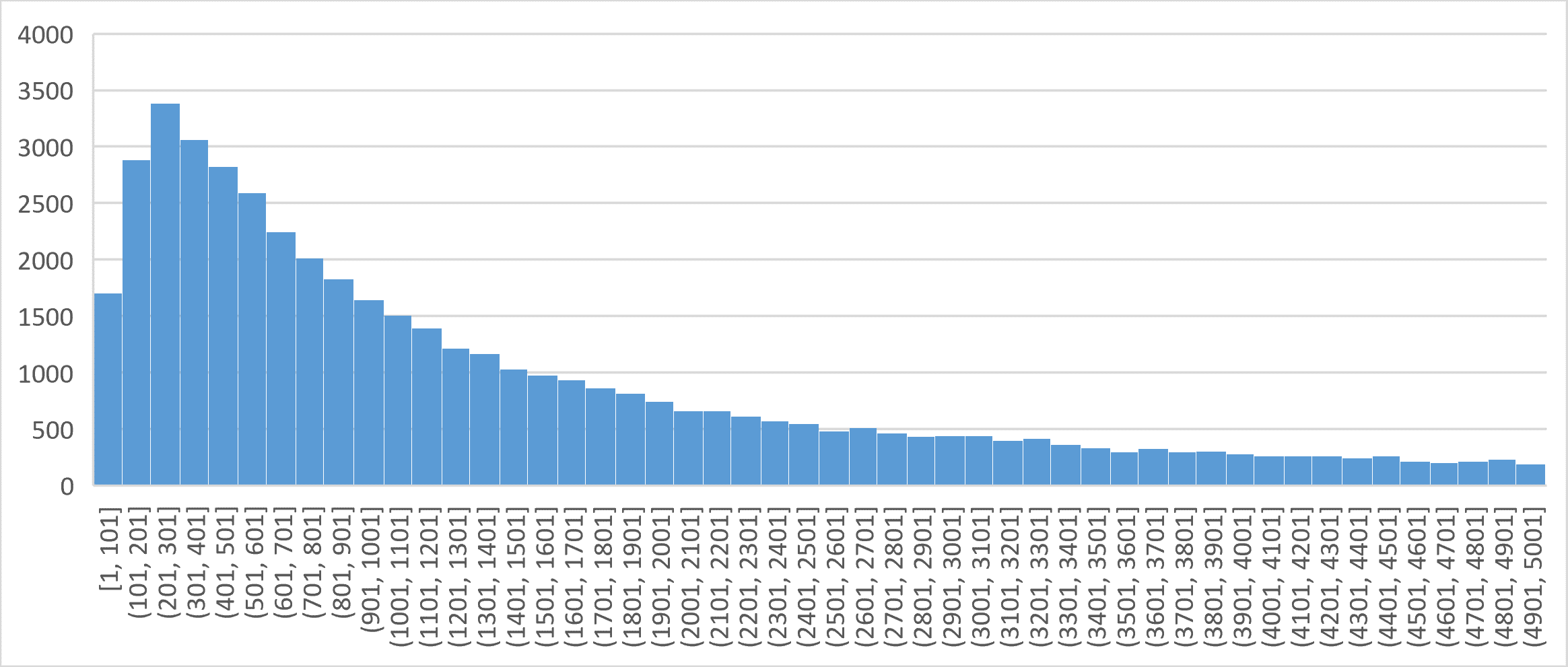} 
        \label{plots:distr}
    }
	\caption{\textbf{Statistical Summary of the created dataset}}
	\label{fig:set_sum}
\end{figure*}

\bfsection{Data Labeling}
After careful examination, we have eliminated redundant images and those without aphid clusters, resulting in a total of 5,447 labeled photos. Redundant images refer to those captured from very close viewpoints, leading to significant visual similarities among the images. This data selection process ensures that the final dataset comprises visually distinct images with the presence of aphid clusters, enabling deep CNN models to learn and generalize effectively from the available data.
In summary, the percentages of photos from views 1, 2, and 3 are shown in Fig~\ref{plots:compo}. 
The size of the labeled masks ranges from 1 to 1,250,193 pixels. 77.0\% of the masks have a size smaller than 5,000 pixels. Since masks with larger sizes are rare and sparsely distributed, we only plot the histogram of masks with less than 5,000 pixels in Fig~\ref{plots:distr}. More than half of the masks are smaller than 1,500 pixels, with the most popular size interval [201, 301]. Among all masks, the median size is 1,442 pixels and the mean is 7,867 pixels. The median is more representative, while the mean value is affected by the extremely large masks.

\bfsection{10-Fold Cross-Validation}
Cross-validation \cite{stone1974cross} is a resampling method to evaluate and pick models on a small dataset. Popular computer vision datasets commonly have more than 10k images, \eg MS COCO \cite{lin2014microsoft} has more than 200k labeled images. Our dataset only has a little over 5,000 images. Following cross-validation \cite{stone1974cross}, we split our dataset into 10 groups. To ensure each group has a similar percentage of images from views 1, 2, and 3, we separately shuffle the images and split them into 10 subgroups for each view. Then the final cross-validation groups are formed by picking one subgroup from each view. Thus, images from each view will be evenly distributed in each group. 

\bfsection{Image Patches}
The majority of the masks, as shown in Fig~\ref{plots:distr}, have a size smaller than 1,500 pixels, which is less than 0.015\% of the original image size ($3,648 \times 2,736$). In addition, most detection and segmentation models are trained and tested on much smaller images. We decide to split the original high-definition images into smaller square patches with side 400 pixels. During this process, some masks will be separated into different patches and will have some exclusions. To ensure each mask's completeness in at least one of the final patches, the patch generation is done with 50\% overlapping, meaning the next patch overlaps 50\% with the previous patch both horizontally and vertically. An original $3,648 \times 2,736$ image generates 221 patches for the detection and segmentation tasks. 

Patch generation is conducted after dividing the dataset into 10 cross-validation groups. This division ensures that no information from one original photo is used in any other group, maintaining the integrity of the evaluation process.
After cropping, we discard patches without an aphid cluster since the CNN models already have sufficient negative samples available from the backgrounds of other patches. By discarding patches without aphid clusters, we can focus on training the models using patches that contain the desired positive samples. In summary, the number of patches in each cross-validation group is shown in Table \ref{table:num_patch} with a total of 151,380 image patches in the dataset.

\begin{table*}
\centering
\caption{\textbf{Number of Patches in Each Group}}
\begin{tabular}{ | c | c | c | c | c | c | c | c | c | c | c | }
    \hline
	Sum  &  1  &  2  &  3  &  4  &  5  &  6  &  7  &  8  &  9  &  10 \\
    \hline
	151,380 & 14,778 & 15,392 & 14,567 & 15,720 & 15,943 & 14,929 & 15,272 & 15,276 & 14,140 & 15,363 \\
    \hline
\end{tabular}
\label{table:num_patch}
\end{table*}

\section{Object Detection Models}

For object detection, the recognition and localization of objects in videos or images require both classification and localization components. Typically, detection models consist of two separate branches dedicated to classification and localization, respectively. The classification branch performs similarly to standard classification tasks, where it categorizes the contents enclosed by bounding boxes. On the other hand, the localization branch predicts the offsets relative to anchor boxes in anchor-based detection models or anchor points in anchor-free detection models. These predicted offsets are then converted into bounding box coordinates based on the anchor boxes or anchor points, contributing to the final predictions.

A crucial factor for building a successful detection model lies in effectively dividing positive and negative samples. Regardless of whether pre-defined anchor boxes or anchor points are utilized, it is necessary to categorize the samples as positives or negatives. Negative samples represent the background and do not contribute to predicting object locations, while positive samples play a vital role in predicting the corresponding object locations. For example, in anchor-based models, anchor samples can be classified as positives if their Intersection over Union (IoU) with ground truth objects exceeds certain thresholds (e.g., 0.5). However, recent detection models have moved away from using fixed IoU thresholds, as small objects may have significantly fewer corresponding positive samples compared to larger objects. Consequently, modern detection models tend to calculate adaptive thresholds based on statistical properties among the samples \cite{zhang2020bridging}\cite{zhang2021varifocalnet} or compute the dynamic thresholds based on the training status \cite{kim2020probabilistic}\cite{zhang2022dynamic}.

\begin{figure*}[ht]
\centering
\includegraphics[width=\linewidth]{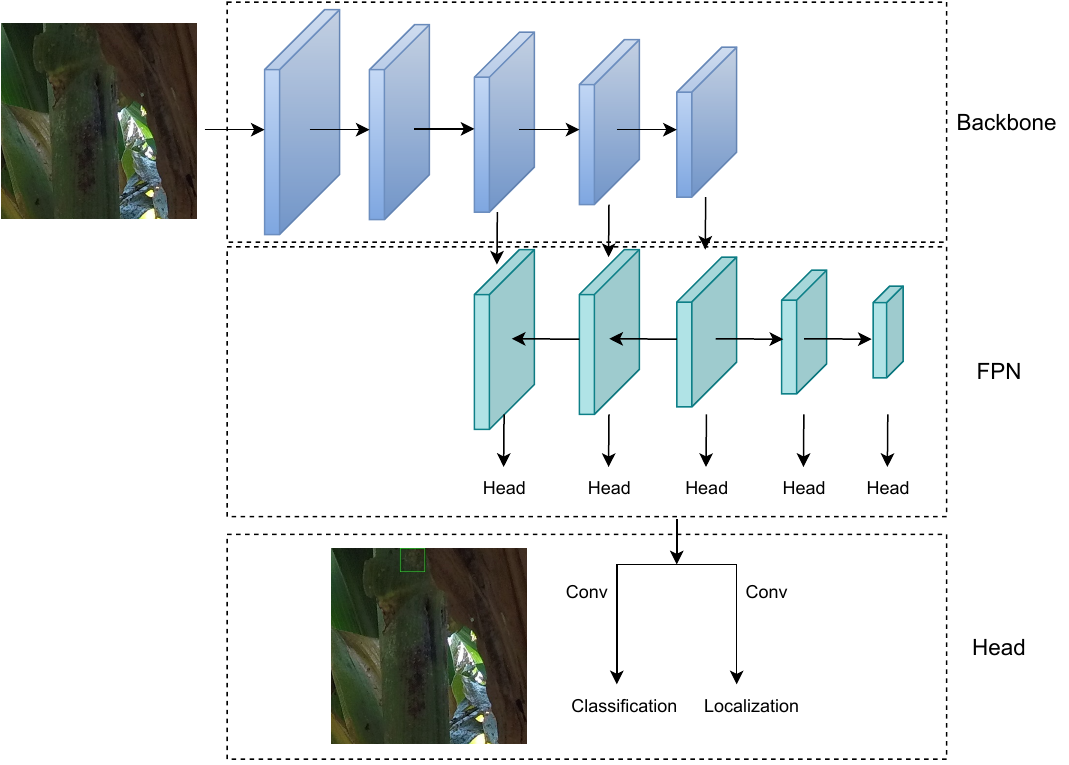}
\caption{A classic detection model with Feature Pyramid Networks (FPN). The backbone is utilized as the feature extractor that acquires the semantic information from the information. FPN propagates the semantic information from the deeper layers to the shallower layers so that all features in FPN could have enough semantic information. The head generates the predictions from the various features from FPN.}
\label{fig:1}
\end{figure*}

Most generic state-of-the-art detection models are based on Feature Pyramid Networks \cite{lin2017feature} where the detection results come from various feature maps with different sizes to detect objects with various shapes and sizes, as illustrated in Figure~\ref{fig:1}. The backbone is Convolutional Neural Networks (CNNs) for extracting the features from the images. Deep layers have relatively strong semantic information while the shallow layer may not contain enough semantic information. Thus, FPN is utilized to merge the adjacent features from the deeper layers to the shallower layers so that the shallow layers could have strong semantic information and be exploited to generate the classification scores and localization predictions. The head is to predict the bounding boxes and classes in the corresponding boxes and the same head is applied to all feature maps from FPN. There are two branches in the head for classification and bounding box regression. Since the NMS (Non-Maximum Suppression) algorithm is based on the classification scores to eliminate the duplicates, some high-quality bounding boxes might be removed due to their low classification scores. Thus some methods estimate localization quality for classification prediction \cite{li2021generalized} or introduce the predictions to divide the positives and negatives dynamically to reduce the gap between the classification branch and regression branch. Recently, Transformers \cite{vaswani2017attention} were introduced to computer vision and some transformer-based detection models \cite{carion2020end}\cite{ma2021miti} achieve excellent results on COCO benchmark \cite{lin2014microsoft}. However, Transformer-based models usually require a much longer time to converge compared to CNN-based models.

We have chosen four state-of-the-art object detectors to train, evaluate, and compare their performance on the aphid dataset that we have created. (1) ATSS (Adaptive Training Sample Selection) \cite{zhang2020bridging} calculates the adaptive IoU thresholds based on the mean and standard deviation of the IoUs between the candidate anchor boxes and the ground truth objects to select the positive samples instead of using fixed thresholds. (2) GFLV2 (Generalized Focal Loss V2) \cite{li2021generalized} utilizes statistics of bounding box distributions as the Localization Quality Estimation (LQE). Thus the high-quality bounding boxes could have a high probability to be kept instead of suppressed with the NMS (Non-Maximum Suppression) algorithm. (3) PAA \cite{kim2020probabilistic} dynamically divides the positive samples and negative samples using GMM (Gaussian Mixture Model) based on the classification and localization scores of the samples in a probabilistic way. (4) VFNet \cite{zhang2021varifocalnet} is based on ATSS \cite{zhang2020bridging} algorithm, and employs IoU-aware Classification Score (IACS) as the classification soft target using the IoUs between the predicted bounding boxes and their corresponding ground truth objects. Thus high-quality predicted boundary boxes might have high scores than those low-quality boxes. In addition, star-shaped box feature representation is introduced to further refine the predicted boxes so that they could be closer to the ground truth objects.

\begin{table*}[ht!]
\small
\centering
\caption{The Mean and Standard Deviation of 10-fold Cross-Validation on State-Of-The-Art Detection Models}
\setlength{\tabcolsep}{18pt}
\renewcommand{\arraystretch}{1.5}
\begin{tabular}{c|c|c|c|c|c}
 \hline
AP (Recall) & Method & VFNet & GFLV2 & PAA & ATSS \\
\hline
\multirow{3}{*}{Test 10} & original & 42.7 (79.3) & 42.4 (78.2) & 42.2 (82.9) & 42.2 (78.6) \\

& +merge 10 & 46.7 (83.9) & 45.9 (82.3) & 45.4 (87.3) & 46.3 (83.0) \\

& +rm 0.01 & 60.3 (97.1) & 60.3 (96.3) & 60.3 (98.5) & 60.6 (97.2) \\
\hline

\multirow{3}{*}{Test 9} & original & 44.2 (81.9) & 44.2 (81.2) & 43.4 (85.3) & 43.9 (81.7) \\

& +merge 10 & 47.1 (85.3) & 47.2 (84.7) & 46.6 (89.0) & 47.1 (85.1) \\

& +rm 0.01 & 60.5 (97.2) & 60.4 (97.1) & 61.4 (98.4) & 61.3 (97.5) \\
\hline

\multirow{3}{*}{Test 8} & original & 42.3 (80.7) & 41.7 (79.6) & 41.4 (84.8) & 42.1 (80.4) \\

& +merge 10 & 45.2 (84.3) & 45.6 (83.3) & 44.6 (87.8) & 45.6 (83.5) \\

& +rm 0.01 & 59.5 (97.2) & 59.1 (96.4) & 58.9 (98.7) & 59.6 (97.2) \\
\hline

\multirow{3}{*}{Test 7} & original & 43.0 (81.3) & 42.7 (80.1) & 42.2 (84.5) & 43.0 (80.7) \\

& +merge 10 & 45.9 (84.3) & 46.1 (83.5) & 45.2 (88.4) & 45.9 (84.7) \\

& +rm 0.01 & 58.0 (96.7) & 58.5 (96.3) & 59.5 (98.5) & 59.6 (97.2) \\
\hline

\multirow{3}{*}{Test 6} & original & 41.4 (80.8) & 41.0 (79.5) & 40.7 (84.6) & 41.5 (80.4) \\

& +merge 10 & 44.2 (83.5) & 43.9 (82.0) & 43.6 (87.9) & 44.0 (83.2) \\

& +rm 0.01 & 57.2 (96.9) & 57.1 (96.1) & 57.8 (98.3) & 57.1 (97.1) \\
\hline

\multirow{3}{*}{Test 5} & original & 43.3 (80.5) & 42.5 (79.6) & 42.4 (83.7) & 43.0 (80.0) \\

& +merge 10 & 46.3 (84.0) & 45.9 (83.3) & 45.4 (87.5) & 46.1 (83.5) \\

& +rm 0.01 & 60.3 (96.3) & 60.6 (95.9) & 60.8 (98.4) & 60.8 (96.8) \\
\hline

\multirow{3}{*}{Test 4} & original & 43.3 (81.0) & 42.9 (79.8) & 42.0 (84.5) & 43.1 (80.7) \\

& +merge 10 & 45.0 (83.8) & 45.4 (82.3) & 44.5 (87.4) & 45.6 (83.6) \\

& +rm 0.01 & 59.0 (96.7) & 59.1 (95.8) & 59.0 (98.4) & 59.8 (96.9) \\
\hline

\multirow{3}{*}{Test 3} & original & 38.9 (79.9) & 38.9 (78.2) & 39.1 (84.1) & 39.5 (79.4) \\

& +merge 10 & 42.2 (83.1) & 41.8 (81.9) & 42.1 (87.3) & 42.5 (82.6) \\

& +rm 0.01 & 56.5 (96.4) & 56.3 (96.0) & 56.5 (97.9) & 57.0 (96.7) \\
\hline

\multirow{3}{*}{Test 2} & original & 41.5 (80.0) & 41.6 (78.8) & 40.9 (83.7) & 41.2 (79.5) \\

& +merge 10 & 44.3 (82.9) & 44.1 (81.6) & 43.5 (87.4) & 43.8 (82.3) \\

& +rm 0.01 & 56.7 (97.2) & 56.6 (96.4) & 57.0 (98.5) & 57.6 (97.2) \\
\hline

\multirow{3}{*}{Test 1} & original & 38.4 (79.0) & 38.2 (76.6) & 37.9 (82.4) & 38.5 (78.5) \\

& +merge 10 & 41.3 (81.9) & 41.3 (81.0) & 41.0 (85.7) & 41.3 (81.7) \\

& +rm 0.01 & 55.2 (96.5) & 55.3 (95.4) & 55.8 (98.1) & 56.2 (96.6) \\
\hline\hline

\multirow{3}{*}{Mean} & original & 41.9 (80.4) & 41.6 (79.2) & 41.2 (84.1) & 41.8 (80.0) \\

& +merge 10 & 44.8 (83.7) & 44.7 (82.6) & 44.2 (87.6) & 44.8 (83.3) \\

& +rm 0.01 & 58.3 (96.8) & 58.3 (96.2) & 58.7 (98.4) & 59.0 (97.0) \\
\hline

\multirow{3}{*}{Std} & original & 1.91 (0.90) & 1.84 (1.27) & 1.65 (0.89) & 1.70 (1.00) \\

& +merge 10 & 1.89 (0.92) & 1.93 (1.09) & 1.68 (0.86) & 1.85 (1.03) \\

& +rm 0.01 & 1.87 (0.35) & 1.90 (0.45) & 1.89 (0.23) & 1.82 (0.28) \\
\hline

\end{tabular}
\label{table:1}
\end{table*}

\begin{table*}[ht!]
\small
\centering
\caption{The Experimental Results with Different IoU Thresholds for Split 1 as the Test}
\setlength{\tabcolsep}{18pt}
\renewcommand{\arraystretch}{1.5}
\begin{tabular}{c|c|c|c|c|c}
 \hline
AP (Recall) & IoU Threshold & VFNet & GFLV2 & PAA & ATSS \\
\hline

\multirow{3}{*}{Original} & 0.5 & 38.4 (79.0) & 38.2 (76.6) & 37.9 (82.4) & 38.5 (78.5) \\

& 0.25 & 51.2 (89.0) & 51.3 (88.2) & 49.2 (91.6) & 51.5 (89.3) \\

& 0.75 & 16.2 (37.0) & 15.5 (34.5) & 16.5 (38.8) & 15.7 (35.7) \\
\hline

\multirow{3}{*}{Merge 10} & 0.5 & 41.3 (81.9) & 41.3 (81.0) & 41.0 (85.7) & 41.3 (81.7) \\

& 0.25 & 54.0 (90.6) & 54.1 (90.1) & 52.5 (93.5) & 54.5 (90.8) \\

& 0.75 & 18.8 (43.4) & 17.8 (40.7) & 19.4 (45.5) & 17.8 (41.6) \\
\hline

\multirow{3}{*}{Merge 10 + rm 0.01} & 0.5 & 55.2 (96.5) & 55.3 (95.4) & 55.8 (98.1) & 56.2 (96.6) \\

& 0.25 & 68.9 (99.0) & 69.1 (98.8) & 68.6 (99.8) & 70.0 (99.2) \\

& 0.75 & 27.5 (61.8) & 26.2 (59.1) & 28.1 (63.5) & 27.3 (61.1) \\
\hline

\end{tabular}
\label{table:2}
\end{table*}

\subsection{Model Training}

The images we collected are real-world images in the fields with high resolutions (i.e., $3,648 \times 2,736$) and various illuminations and occlusions, making it extremely challenging to detect aphids using existing detection models, even when the bounding boxes are based on aphid clusters rather than individual aphids. To address this challenge, we perform cropping on each labeled image, dividing them into smaller patches of size $400\times400$ for more effective detection. Additionally, the entire image patch dataset is divided into 10 subsets and we evaluate the detection models on the dataset by 10-fold cross-validation.

All models exploit 0.001 as the initial learning rate with the total training epoch being 12. The initial learning rate is utilized for 9 epochs and the learning rate is reduced by 10 for the last 3 epochs. SGD (Stochastic Gradient Descent) is employed as the optimizer to train the model. The momentum and weight decay are 0.9 and 0.0005, respectively. The batch size is 16 and the warmup iterations are 500. The detection models are implemented using PyTorch with Python3 \cite{chen2019mmdetection}.

The evaluation metric used for the detection models is Average Precision (AP), which measures the area under the Precision-Recall (PR) curve. The PR curve illustrates the trade-off between the Precision rate and the Recall rate of the detection models. The Precision rate quantifies the proportion of correctly predicted samples among all predicted positive samples. It is calculated as the ratio of the number of true positive samples to the total number of predicted positive samples. The Recall rate represents the proportion of correctly predicted samples among all ground truth positive samples. It is computed as the ratio of the number of true positive samples to the total number of actual positive samples. The precision and the recall are calculated below.

\begin{equation} \label{eq:1}
Precision = \frac{True\ Positive}{True\ Positive + False\ Positive}
\end{equation}

\begin{equation} \label{eq:2}
Recall = \frac{True\ Positive}{True\ Positive + False\ Negative}
\end{equation}

In object detection tasks, it is essential to not only predict the correct labels but also consider the accuracy of the bounding boxes. The quality of the predicted bounding boxes is commonly evaluated using IoU (Intersection over Union), which measures the overlap between predicted bounding boxes and their corresponding ground truth boxes. IoU is calculated as the ratio of the intersection area to the union area of two bounding boxes.
PASCAL VOC \cite{everingham2010pascal} selects 0.5 as the IoU threshold which indicates that the detection is a success if the IoU between the predicted bounding box and the ground truth bounding box is over 0.5 if the classification label is correctly predicted. COCO \cite{lin2014microsoft} chooses the IoU threshold from 0.5 to 0.95 with the step of 0.05, calculates the AP for each of the thresholds, and finally averages them. In this paper, we harness the IoU threshold from PASCAL VOC and the generated annotation files are also in xml format used by PASCAL VOC \cite{everingham2010pascal}.

\begin{figure*}[htp]
\centering
\includegraphics[width=.24\textwidth]{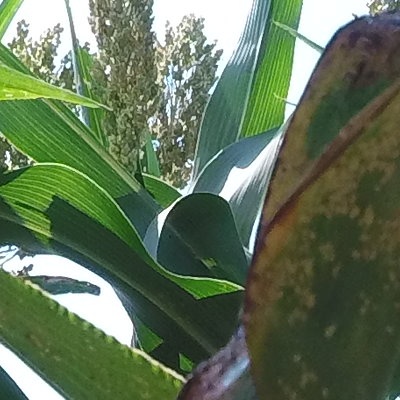}
\includegraphics[width=.24\textwidth]{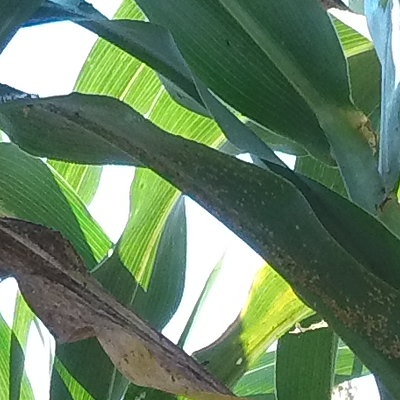}
\includegraphics[width=.24\textwidth]{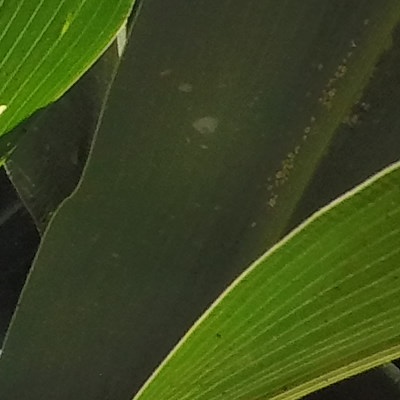}
\includegraphics[width=.24\textwidth]{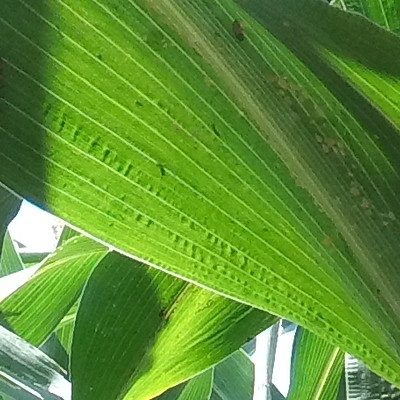}\\
\includegraphics[width=.24\textwidth]{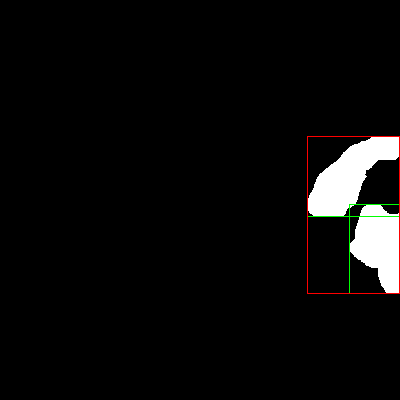}
\includegraphics[width=.24\textwidth]{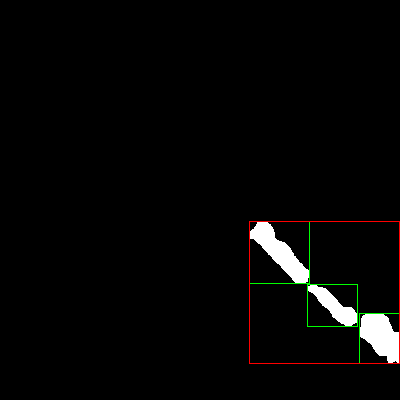}
\includegraphics[width=.24\textwidth]{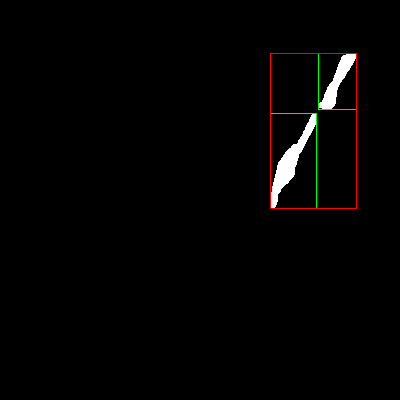}
\includegraphics[width=.24\textwidth]{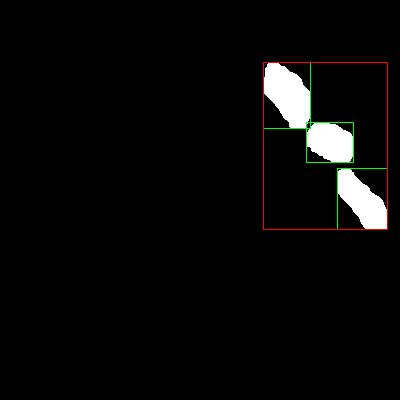}
\caption{Some examples of merging clusters that are close enough (10 pixels) to each other. The segmentation masks are utilized to illustrate the changes in the bounding boxes before and after the merging process. The white areas are the masks for aphid clusters. The green bounding boxes in the masks are the originally labeled boxes and the red bounding boxes in the masks are the merged results. Some clusters are merged as a whole for detection and the detection models might not recognize those small clusters individually but could be relatively easy to detect as a whole.}
\label{fig:2}
\end{figure*}

\begin{figure*}[htp]
\centering
\includegraphics[width=.24\textwidth]{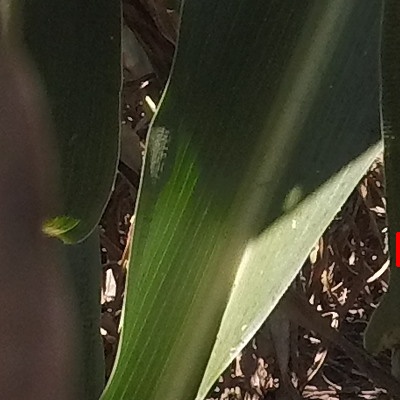}
\includegraphics[width=.24\textwidth]{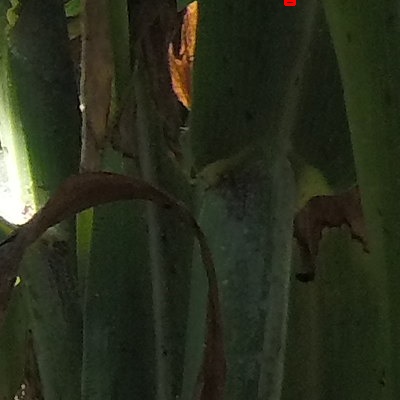}
\includegraphics[width=.24\textwidth]{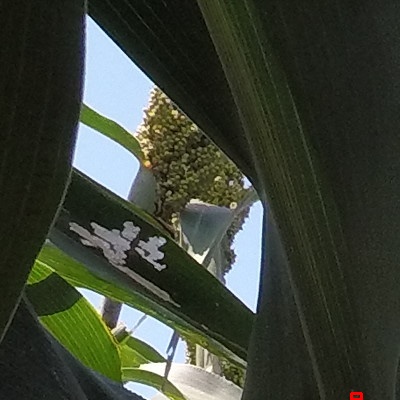}
\includegraphics[width=.24\textwidth]{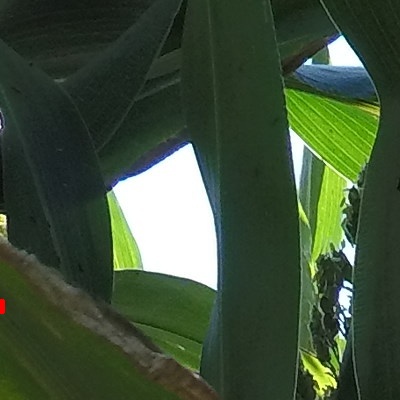} \\
\caption{Several examples of image patches contain extremely small partial ground truth after cropping. In most cases, the small partial ground truth bounding boxes appear at the border of the image patches, which is illustrated above with red bounding boxes. If those image patches are in the training set, the contribution of those patches is almost zero for training the detection models. If those image patches are in the testing set, the overall performance would drop since even the best detection model cannot accurately localize those small partial aphid clusters at the border. Thus removing those small partial aphid clusters is meaningful in both training and evaluation.}
\label{fig:3}
\end{figure*}

\section{Results}
During the experiments, we implemented and compared four state-of-the-art detection models using a 10-fold cross-validation approach. To facilitate the evaluation process, we have cropped $400\times400$ patches from the high-resolution images and organized them into 10 distinct groups for cross-validation. In each fold of the cross-validation, one group is designated as the testing data while the remaining groups are merged to form the training data for that specific validation. However, due to the nature of labeling aphids based on clusters, some small clusters that are in close proximity to each other are labeled individually, as shown in Figure~\ref{fig:2}. In this case, the bounding boxes overlap, which may confuse the learning models during training and affect the performance of the detection models.

To address this issue, we preprocess the ground truth labels by merging the bounding boxes of small clusters if they are sufficiently close to each other. Specifically, in our experiments, we merge the bounding boxes of clusters if their closest distance is less than or equal to 10 pixels, as illustrated in Figure~\ref{fig:2}. By merging these closely located clusters, we treat them as a whole object, enabling the detection models to more effectively recognize and detect these merged clusters. Our experiments demonstrate that this strategy is particularly useful for small clusters that are challenging to detect, even with the utilization of state-of-the-art detectors. Moreover, in cases where clusters are densely located and difficult to separate, the merging process facilitates the detection models and boosts their performance in real-world scenarios.


When we crop the original images into $400\times400$ small patches, some aphid clusters are cropped into different patches that may result in extremely small patches along the border of the image patches. These tiny patches contribute nothing to the training process but greatly compromise the accuracy of the testing data. Some examples of the image patches that contain the tiny partial aphid clusters after cropping is demonstrated in Figure~\ref{fig:3}. These tiny cluster patches are located along the border of the image patches. To further improve the model performance, we  remove these tiny clusters (i.e., areas that are less than 1\% of the patch) from the annotation. As demonstrated in Table~\ref{table:1}, the performance of all models is improved by around 17\% after removing these tiny clusters.

In Table~\ref{table:1}, "Test 9" indicates folder 9 is utilized as the testing data and the others are merged for training. Since 10-fold cross-validation is performed, we conducted 10 independent experiments for each fold validation. The average precision (AP) and Recall of each experiment are tabulated in Table~\ref{table:1} for different models. The mean and the standard deviation are calculated across the 10-fold cross-validation. In the "Method" column, ``original" indicates the detection models are applied to the originally labeled dataset; ``+merge 10" indicates the results after merging the bounding boxes of close clusters. ``+rm 0.01" represents the results after removing the tiny clusters.



Based on the observations from Table~\ref{table:1}, it is evident that all the detection models exhibit similar performance in terms of both Average Precision (AP) and recall across different versions of the datasets (original, +merge 10, and +rm0.01). Analyzing the mean and standard deviation values provided in Table~\ref{table:1}, we can conclude that all four models yield comparable results, although the recall rate of the PAA \cite{kim2020probabilistic} model is slightly higher compared to the other detectors. This is primarily due to the fact that PAA generates a larger number of predicted bounding boxes compared to the other models. For instance, in the case of the ``+merge 10" version of the dataset, the mean recall for PAA is 87.6, while the corresponding values for the other three models are 83.7, 82.6, and 83.3, respectively. Additionally, examining the standard deviation values across all validations, we can observe that the variations in AP and recall are small for all the detection models. This suggests that the detection results are consistent and stable across each validation, further bolstering the reliability of the findings.

To assess the impact of IoU thresholds on the experimental results, we have conducted additional experiments with varying IoU thresholds. We used split 1 as the test set and the remaining 9 splits were combined to form the train set in this experiment. The IoU thresholds represent the required IoU value between the predicted bounding box and the ground truth bounding box for successful detection. For example, an IoU threshold of 0.5 indicates that a prediction would be considered successful if its IoU with the corresponding ground truth is greater than 0.5, along with the correct classification. Therefore, lowering the IoU threshold generally leads to improved performance of the detection models, and vice versa.

Table~\ref{table:2} shows the experimental results. It clearly demonstrates that as the IoU threshold decreases from 0.5 to 0.25, the accuracy of the detection models increases. Conversely, increasing the IoU threshold from 0.5 to 0.75 results in a decrease in accuracy. This trend highlights the trade-off between strict localization requirements and overall detection accuracy. If the precise location of aphid clusters is not of paramount importance, a lower IoU threshold, such as 0.25, can be employed. This threshold allows for a more lenient criterion in determining successful detections while still maintaining a satisfactory level of accuracy.

\begin{figure*}[htp]
\centering
\includegraphics[width=.24\textwidth]{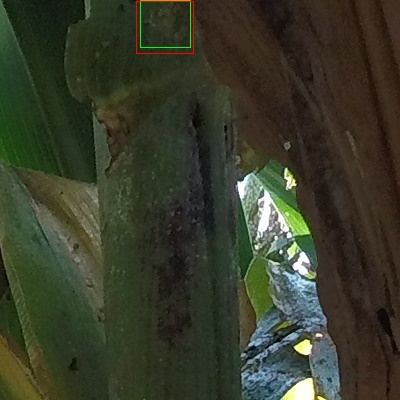}
\includegraphics[width=.24\textwidth]{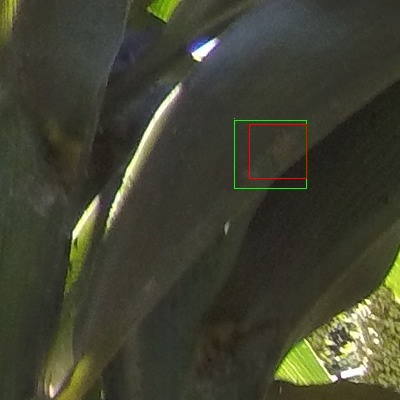}
\includegraphics[width=.24\textwidth]{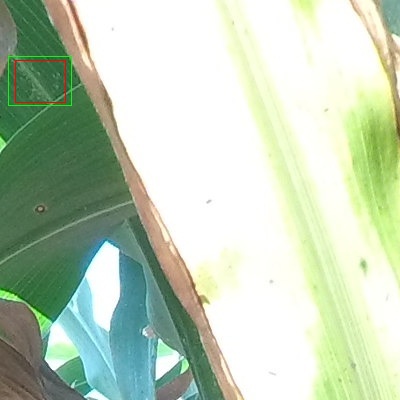}
\includegraphics[width=.24\textwidth]{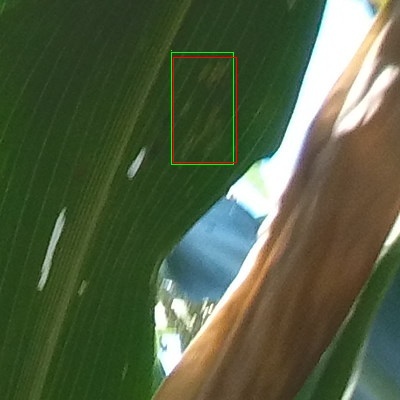} \\
\caption{Some good detection results with VFNet \cite{zhang2021varifocalnet} as the detection model. The red bounding boxes are the ground truth boxes and the green bounding boxes are the predicted bounding boxes. The predicted bounding boxes whose confidence scores are higher than 0.3 are shown in the image patches. Those detection results are almost perfect considering the minor errors when those ground truth clusters are labeled.}
\label{fig:4}
\end{figure*}

\begin{figure*}[ht]
\centering
\includegraphics[width=.24\textwidth]{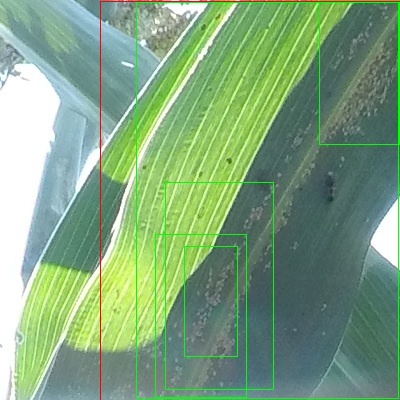}
\includegraphics[width=.24\textwidth]{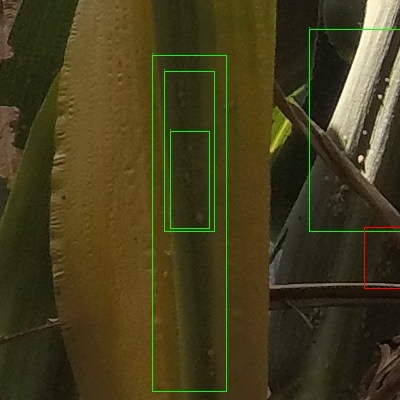}
\includegraphics[width=.24\textwidth]{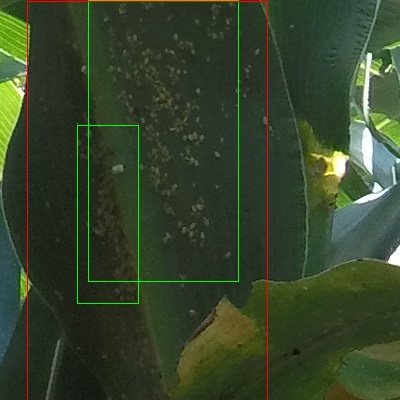}
\includegraphics[width=.24\textwidth]{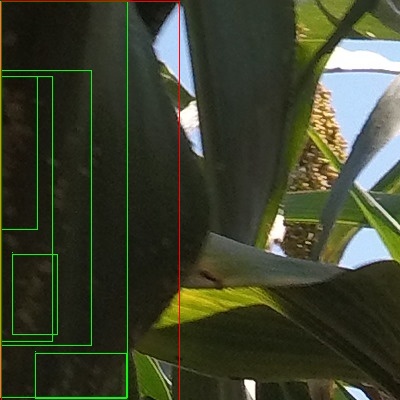} \\
\caption{Some bad detection results with VFNet \cite{zhang2021varifocalnet} as the detection model. The red boxes are the ground truth and the green boxes are the predicted bounding boxes. The predicted bounding boxes whose confidence scores are higher than 0.3 are shown in the image patches. The most common problem for detecting aphid clusters is that there are many duplicates generated around the ground truth clusters (false positives) since there are no fixed sizes and shapes for aphid clusters, as illustrated in the examples above. Thus the predicted bounding boxes for parts of the clusters are frequently generated.}
\label{fig:5}
\end{figure*}

\section{Discussion}


In recent years, the application of deep learning models for object detection and recognition in real-world scenarios has gained significant popularity and success. However, when it comes to recognizing and localizing insects like aphids, several challenges arise due to their small size and the inherent difficulties in capturing high-quality images in real-world agricultural settings. Moreover, from an agricultural perspective, it is not practical or meaningful to detect individual aphids if the objective is to address infestations or mitigate potential damage. Instead, focusing on detecting aphids as clusters is more applicable and relevant. Thus, it is desirable to identify and label aphids as clusters. This approach recognizes the natural tendency of aphids to cluster together, allowing for a more practical and effective detection strategy in real-world agricultural contexts.


Labeling aphid clusters presents a unique challenge due to the absence of well-defined boundaries compared to common objects like cars or humans. Unlike such objects, aphid clusters exhibit irregular shapes and sizes that can vary significantly from one cluster to another. To address this challenge, we employ a two-step labeling approach: first, we label the aphid clusters using masks, and then we generate bounding boxes based on these masks, as depicted in Figure~\ref{fig:2}. The irregular shapes and sizes of the aphid cluster masks make their detection more challenging compared to objects with regular shapes and sizes. However, the area covered by these masks or bounding boxes can provide valuable insights into the severity of aphid infestation. Larger areas occupied by aphid clusters indicate a more severe infection, signaling the need for protective measures and appropriate interventions. 

\begin{figure*}[ht]
\centering
\includegraphics[width=.33\textwidth]{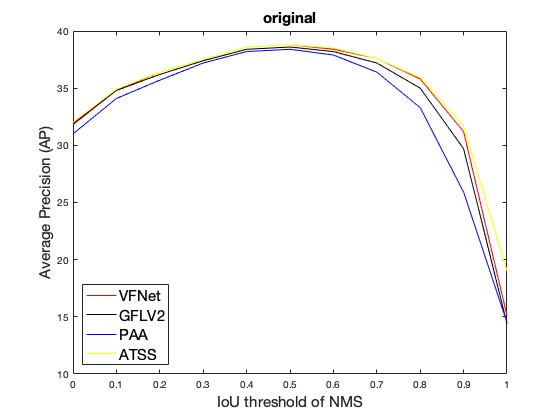}
\includegraphics[width=.33\textwidth]{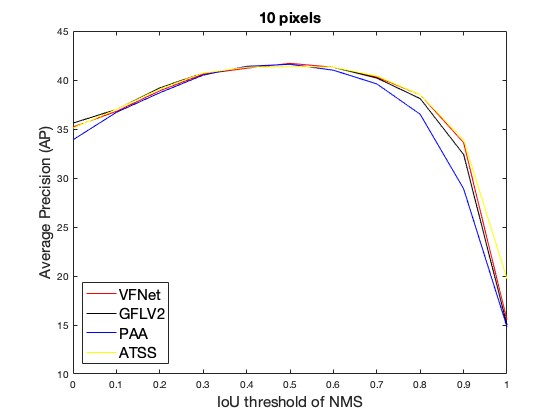}
\includegraphics[width=.33\textwidth]{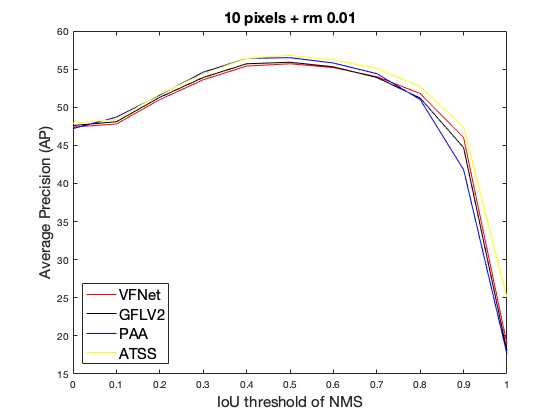}
\caption{The IoU threshold in NMS algorithm v.s. average precision. Similar to Table~\ref{table:1}, ``original" indicates the dataset is our originally labeled dataset; ``10 pixels" represents the neighboring clusters within 10 pixels are merged with a single bounding box; ``10 pixels + rm 0.01" illustrates that merging neighboring clusters within 10 pixels and removing tiny clusters (less than 0.01 of the area of the patches) are implemented for the dataset. Split 1 is utilized as the testing data and the other 9 splits are merged as the training data. The four detection models have the same trend and the performance is extremely unpleasing when the threshold is close to 1. The IoU threshold of around 0.5 could yield the best performance with all detection models and annotations. In the experiments, 0.6 is utilized as the default IoU threshold for NMS algorithm and all the experimental results are based on 0.6 as the IoU threshold of NMS.}
\label{fig:6}
\end{figure*}


Due to the irregular shapes and sizes of aphid clusters, the initial labeling process may not yield perfect accuracy, necessitating post-processing techniques. A straightforward and effective method to enhance detection performance is to merge the bounding boxes of small, neighboring clusters. This approach is particularly beneficial as many state-of-the-art object detectors struggle to accurately recognize small objects. As demonstrated in Table~\ref{table:1}, merging the bounding boxes of neighboring clusters leads to notable improvements in detection performance. The performance is further improved after removing the tiny clusters caused by image cropping. Using the generated dataset, state-of-the-art detection models can be directly employed to detect and recognize the aphid clusters.

In our experiments, all four models yield comparable results. Some detection results are visualized in Figure~\ref{fig:4} using VFNet \cite{zhang2021varifocalnet} as the detection model. We can see that the detection results are very close to the ground truth.  However, the irregular sizes and shapes of aphid clusters pose a challenge, occasionally leading to duplicate detections around the true clusters. As a consequence, the number of false positives increases, thereby impacting the overall accuracy of the detection. Figure~\ref{fig:5} showcases some instances of poor detection results, highlighting the difficulty in accurately detecting irregular aphid clusters. Additionally, it is important to note that certain sorghum diseases, such as leaf blight, can result in leaf lesions that bear similarities to aphid clusters. As a consequence, the models may erroneously identify the lesions as aphids, leading to false positive detections. This can be mitigated by incorporating a broader range of training samples to provide the models with a more comprehensive understanding of the various appearances and features. 

The NMS (Non-Maximum Suppression) algorithm is commonly employed in detection models to eliminate duplicate detections. It operates by setting an IoU threshold, which determines the overlap allowed between bounding boxes before considering them duplicates. The NMS algorithm ranks the detection results by their confidence scores, starting with the highest score. The top-scoring result is retained, while any subsequent results with bounding boxes that have an IoU exceeding the threshold are discarded. 
To explore the influence of IoU thresholds on detection performance, we conduct experiments by varying the IoU thresholds, as depicted in Figure~\ref{fig:6}. The results indicate that the IoU threshold of 0.5 yields the best performance for all detection models and datasets. While varying the IoU threshold in the NMS algorithm does not significantly enhance the performance of the detection models on the aphid dataset.

\section{Conclusion}

In this study, we have collected a large aphid dataset from sorghum fields and annotated the dataset based on aphid clusters. We have also implemented and compared the performance of four state-of-the-art detection models on the dataset and achieved promising results on aphid cluster detection. To improve the detection performance, we preprocessed the dataset by merging the bounding boxes of neighboring clusters and removing extremely small clusters. Consequently, our annotated dataset can be readily utilized with existing object detection models, thus increasing its relevance in real-world scenarios. The resulting dataset and trained models hold the potential to assist farmers in accurately estimating aphid infestation levels and enabling timely and precise pesticide application. Furthermore, the approach and analysis serve as a valuable resource that can inspire further research for the detection and recognition of other insects.


\section{Acknowledgement}
This work was partly supported by the Natural Sciences and
Engineering Research Council of Canada (NSERC) under grant no. RGPIN-2021-04244, and the United States Department of Agriculture (USDA) under Grant no. 2019-67021-28996.

\section{Data availability}
The created dataset is available at Harvard Dataverse via {https://doi.org/10.7910/DVN/N3YJXG}.

\bibliography{sample}

\end{document}